\documentclass{midl} 

\usepackage[T1]{fontenc}    
\usepackage{hyperref}       
\usepackage{url}            
\usepackage{booktabs}       
\usepackage{amsfonts}       
\usepackage{nicefrac}       
\usepackage{microtype}      
\usepackage{float}
\usepackage{amsmath,array,graphicx}
\usepackage{url}
\usepackage{multirow}
\usepackage{hyperref}
\usepackage{todonotes}
\usepackage{subcaption}

\date{August 2018}

\jmlrvolume{-- Under Review}
\jmlryear{2019}
\jmlrworkshop{Full Paper -- MIDL 2019 submission}
\editors{Under Review for MIDL 2019}

\title[Generative Image Translation for Data Augmentation of Bone Lesion Pathology]{Generative Image Translation for Data Augmentation of Bone Lesion Pathology}

\midlauthor{\Name{Anant Gupta\midljointauthortext{Work done when author was at Imagen Technologies.}} \Email{anantgupta@nyu.edu}\\
\addr Courant Institute, New York University \AND
\Name{Srivas Venkatesh} \Email{srivas@imagen.ai}\\
\Name{Sumit Chopra} \Email{sumit@imagen.ai}\\
\Name{Christian Ledig} \Email{christian@imagen.ai}\\
\addr Imagen Technologies, New York }
 
\begin{document}

\maketitle

\begin{abstract}
Insufficient training data and severe class imbalance are often limiting factors when developing machine learning models for the classification of rare diseases. In this work, we address the problem of classifying bone lesions from X-ray images by increasing the small number of positive samples in the training set. We propose a generative data augmentation approach based on a cycle-consistent generative adversarial network that synthesizes bone lesions on images without pathology. We pose the generative task as an image-patch translation problem that we optimize specifically for distinct bones (humerus, tibia, femur). In experimental results, we confirm that the described method mitigates the class imbalance problem in the binary classification task of bone lesion detection. We show that the augmented training sets enable the training of superior classifiers achieving better performance on a held-out test set. Additionally, we demonstrate the feasibility of transfer learning and apply a generative model that was trained on one body part to another.
\end{abstract}


\section{Introduction}

Deep neural networks have demonstrated their potential to reach human-level performance for image classification, however, their performance generally correlates with the amount of available samples \cite{domingos2012few}. When focusing on rare medical conditions, the limited availability of pathological (positive) training images can cause severe class imbalance that limits the accuracy of these models. In contrast, the collection of normal (negative) cases is often substantially simpler. One example of a pathology that is both of high interest but also rare are bone lesions \cite{franchi2012epidemiology}. The classification of the presence of bone lesion pathology in X-ray images is the subject of our work.

Traditional methods to handle class-imbalance, such as image transformations \citep{hussain2017differential} and different sampling strategies \citep{li2010learning, dubey2014analysis}, are often of limited benefit as they do not address the inherent problem of dealing with a small training set not fully representing the underlying data distribution. Recent works have proposed the use of synthetic data in order to augment and increase diversity in the training set \citep{antoniou2017data, mariani2018bagan}. However, learning to generate high-resolution images from random noise requires an often prohibitively large training dataset.

\begin{figure}
\includegraphics[width=13cm]{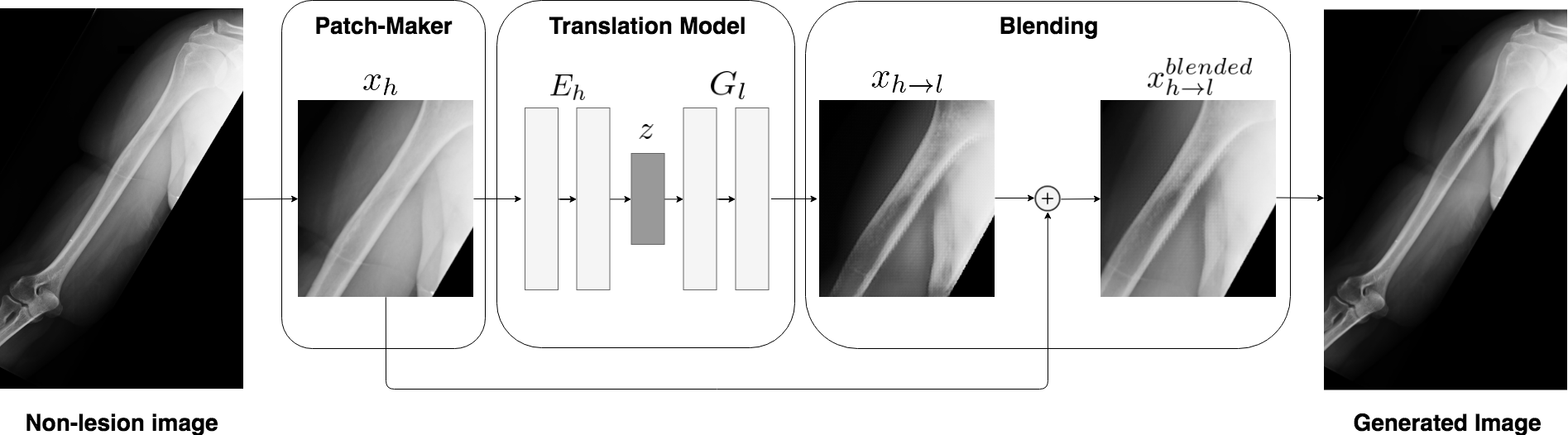}
\centering
\caption{Pipeline of the lesion generation process on non-lesion images. $x_h$ is non-lesion patch; $E_h$, $z$ and $G_l$ are non-lesion encoder, latent representation and lesion generator respectively. $x_{h \rightarrow l}$ is the generated lesion and $x_{h \rightarrow l}^{blended}$ is the result after alpha-blending.}
\label{pipeline}
\end{figure}

In this work, we aim to synthesize bone lesions by translating spatially-constrained patches extracted from non-pathological X-ray images rather than generating from scratch. The lesion-generation pipeline is illustrated in \figurename~\ref{pipeline}. The model is trained on patches to ensure localized generation of pathology. A blending approach merges the translated patches back into those full-images. A subset of the generated images is filtered to form the augmented training set by performing pseudo-labelling. We observed non-trivial performance gains in the task of bone lesion detection for individual body parts (humerus, tibia, femur) when trained using this augmented set. We further show that transfer learning can be a viable option to enhance the training set of body parts for which a powerful image-translation model cannot be trained due to insufficient or noisy samples.

\section{Related Work}
Data augmentation is a well-studied problem in machine learning. Employing transformation-based augmentation techniques \citep{rajkomar2017high, kohli2017medical} or transfer learning by using pretrained weights, are common approaches \citep{rajkomar2017high}, which are used in this work as well.

Generative Adversarial Networks (GANs) \citep{goodfellow2014generative} have been used successfully in the medical imaging domain to accomplish tasks such as image translation \citep{wolterink2017deep, nie2017medical}, segmentation \citep{xue2018segan, kamnitsas2017unsupervised} and data augmentation. \citet{shin2018medical} generate brain tumors for data augmentation by translating segmentation masks to multi-parameteric magnetic resonance (MR) images, using a multi-modal dataset with uniform view. \citet{frid2018gan} use a small dataset of regions of interest of liver lesions in CT images to train a DCGAN \citep{radford2015unsupervised} and generate an augmented set. In comparison, our method focuses on generative data augmentation using a small number of high-resolution X-ray images often varying in positional view even within a single body part.

\citet{salehinejad2017generalization} use DCGAN to generate chest X-rays for multiple pathologies. Plausible samples are filtered out by a team of radiologists to create an augmented set. In our work we perform filtering in an automated manner to mine for hard positives. Recent work by \citet{lau2018scargan} generate scars on cardiac MR scans and employ a blending mask to remove unwanted artifacts. To the best of our knowledge, there is no existing literature that addresses the problem of bone lesion classification by automatically generating pathology in normal radiographs to enhance the training set.

\section{Methodology}
\subsection{Unsupervised Image-to-Image Translation Model}
The generation of bone lesions is posed as an unsupervised image to image translation task \cite{liu2017unsupervised}. In this task, $P{\chi_l}(x_l)$ and $P{\chi_h}(x_h)$ are two marginal distributions from which X-ray patches of bones with lesion $x_l$, and non-lesion $x_h$ are drawn respectively. The model maps these samples to a shared latent representation, using encoders for respective distributions: $E_l(x_l) = E_h(x_h) = z \in \mathcal{Z}$. The generators respective to each distribution decode back the input sample from this latent vector: $G_l(z) = x_l, G_h(z) = x_h$. Lesion-like properties are generated in a normal bone X-ray with the following translation operation: $G_l(E_h(x_h)) = x_{h \rightarrow l}$. This framework is based on the assumption that there exists an unknown but finite joint distribution $P_{\chi_l, \chi_h}$, which the shared latent space can learn to represent.

In order to find the optimal hypothesis for this problem, the lesion encoder-generator is a variational autoencoder (VAE) \cite{kingma2013auto}, whose loss function maximizes the Evidence Lower Bound (ELBO) by minimizing the following objective:
\begin{equation}
\mathcal{L}_{\text{\tiny VAE}_l} = \lambda_1\text{KL}(q_l(z_l|x_l)||\mathcal{N}(z|0, I)) - \lambda_2 \mathbb{E}_{z_l \sim  q_l(z_l|x_l)}[\log p_{G_l}(x_l|z_l)]
\end{equation} 
where $q_l(z_l|x_l)$ is the distribution from which $z_l$ (encoding of $x_l$) is sampled. In the first term, the KL divergence between this distribution and the prior is minimized, which encourages $q_l(z_l|x_l)$ to follow a normal (zero-mean, unit-covariance) distribution. The second term aims to maximize the log-likelihood of $p_G$. The same formulation is followed to train a second VAE for normal, non-lesion samples. This would ensure each generator is able to reconstruct images of the respective distribution.

An adversarial objective \cite{goodfellow2014generative} is employed to help in learning to translate from one domain to another. In this setting, the lesion generator $G_l$ is conditioned on the latent encoding of a healthy patch $z_h$ and the generated sample is evaluated by the discriminator $D_l$ to classify whether the sample was drawn from $P{\chi_l}(x_l)$ or not. This encourages the generator to create lesion-like image features while constructing an image sample. The GAN objective is defined as:
\begin{equation}
\mathcal{L}_{\text{\tiny GAN}_l} = \lambda_0[\mathbb{E}_{x_l \sim P_{\chi_1}}[\log D_l(x_l)] + \mathbb{E}_{z_h \sim q_h(z_h|x_h)}[\log (1 - D_l(G_l(z_h)))]]
\end{equation}

The conceptual shared latent space is implemented in practice by weight-sharing across the two VAEs. The shared latent space implies a cycle-consistency constraint \cite{zhu2017unpaired} that ensures successful circular back and forth mapping between domains:
\begin{align}
\mathcal{L}_{\text{\tiny CC}_l} = &\lambda_3[\text{KL}(q_l(z_l|x_l)||\mathcal{N}(z|0, I)) + \text{KL}(q_h(z_h|x_{l \rightarrow h})||\mathcal{N}(z|0, I))] \nonumber \\
&- \lambda_4 \mathbb{E}_{z_h \sim  q_h(z_h|x_{l \rightarrow h})}[\log p_{G_l}(x_l|z_h)]
\end{align}

This objective aims at preserving the original information of the input image and prevents mode collapse by translating all images to a single output image. Similar loss objectives are minimized for $\text{VAE}_h$, $\text{GAN}_h$ and $\text{CC}_h$. The hyperparameters ($\lambda$) control the contribution of each respective loss function. The network is jointly trained to optimize the following objective:
\begin{align}
\min_{E_l,E_h,G_l,G_h} \max_{D_l,D_h} 
\mathcal{L}_{\text{\tiny VAE}_l} + \mathcal{L}_{\text{\tiny GAN}_l} + \mathcal{L}_{\text{\tiny CC}_l} +
\mathcal{L}_{\text{\tiny VAE}_h} + \mathcal{L}_{\text{\tiny GAN}_h} + \mathcal{L}_{\text{\tiny CC}_h}
\end{align}

\subsection{Patch-making}
\label{subsec:patches}
Bone lesions tend to cause local alterations in bone anatomy without substantially affecting the global visual appearance of the image. We therefore aim to translate localized image patches rather than training a translation model for the complete images. This technique has the following advantages: i) computationally cheaper, ii) multiple patches can be created from a single image, iii) lesion-like features are more prominent on the localized patch, which supports efficient training of the translation model. 
Lesion patches are created by randomly cropping a square area (if image size permits) by a factor $s\in\{1,2\}$ larger than the larger side of the bounding box around the area containing the pathology. This area is marked with a manually annotated bounding box (c.f. \figurename~\ref{fig:patch}). We employ an heuristic to automate cropping of normal patches. We identify potential `crop-areas' in a two step process. First we randomly choose $n$ similar non-lesion images for each lesion image. Second we crop each non-lesion image based on the lesion annotations of the matched lesion image. All non-lesion patches with a mean, normalized ([0,1]) pixel intensity of less than 0.15 are assumed to not contain bone structure and are dropped from the dataset.

\subsection{Blending}
The translated patches also exhibit subtle changes in the overall image characteristics, such as contrast and brightness. This leads to the patch being visibly distinct when placed back in the full-image after translation. We employ alpha-blending to smoothly blend the translated patch in the original image: $x_{h \rightarrow l}^{blended} = \alpha x_{h \rightarrow l} + (1 - \alpha) x_h$. Specifically, we define a locally varying blending factor $\alpha$ as: $\alpha = \cos(|i|^n * \frac{\pi}{2}) \cos(|j|^n * \frac{\pi}{2})$, where $i$ and $j$ are the normalized ([-1,1]) coordinates of a pixel in the patch and $n$ is a hyper-parameter. 

\subsection{Pseudo-Labelling}
We aim to augment the training set with images containing a prominent lesion after the blending operation. We perform hard positive mining \cite{lee2013pseudo} on the generated set using a classifier trained on the available empirical training data (baseline). Based on a threshold parameter $t$, the baseline classifier segregates the generated samples into two disjoint sets: samples with extreme lesion-like properties, and noisy samples. The former is used for augmentation and added to the training set.

\begin{figure}
    \subfigure[Humerus]{\label{fig:humerus}%
      \includegraphics[width=2.5cm,height=3.1cm]{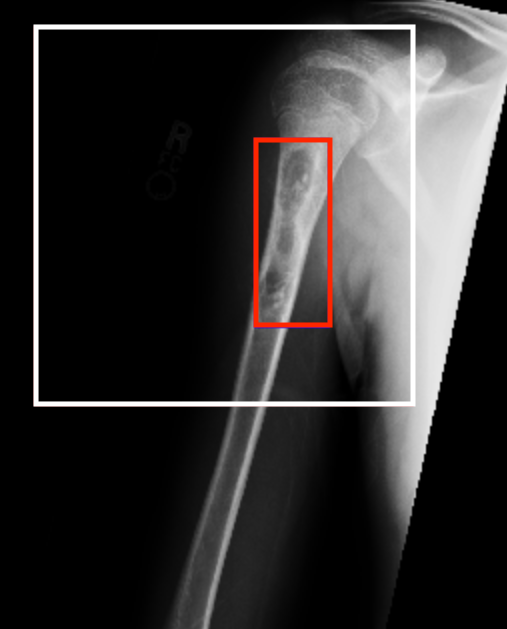}
      \includegraphics[width=2.5cm,height=3.1cm]{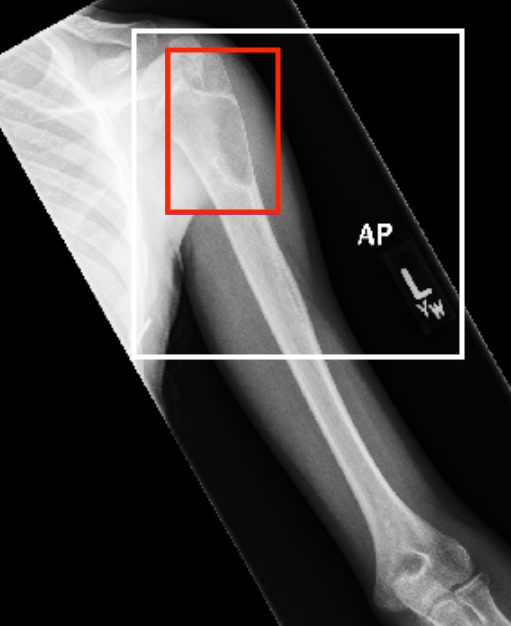}}
      \hspace{1mm}
      \subfigure[Tibia]{\label{fig:tibia}%
      \includegraphics[width=2.5cm,height=3.1cm]{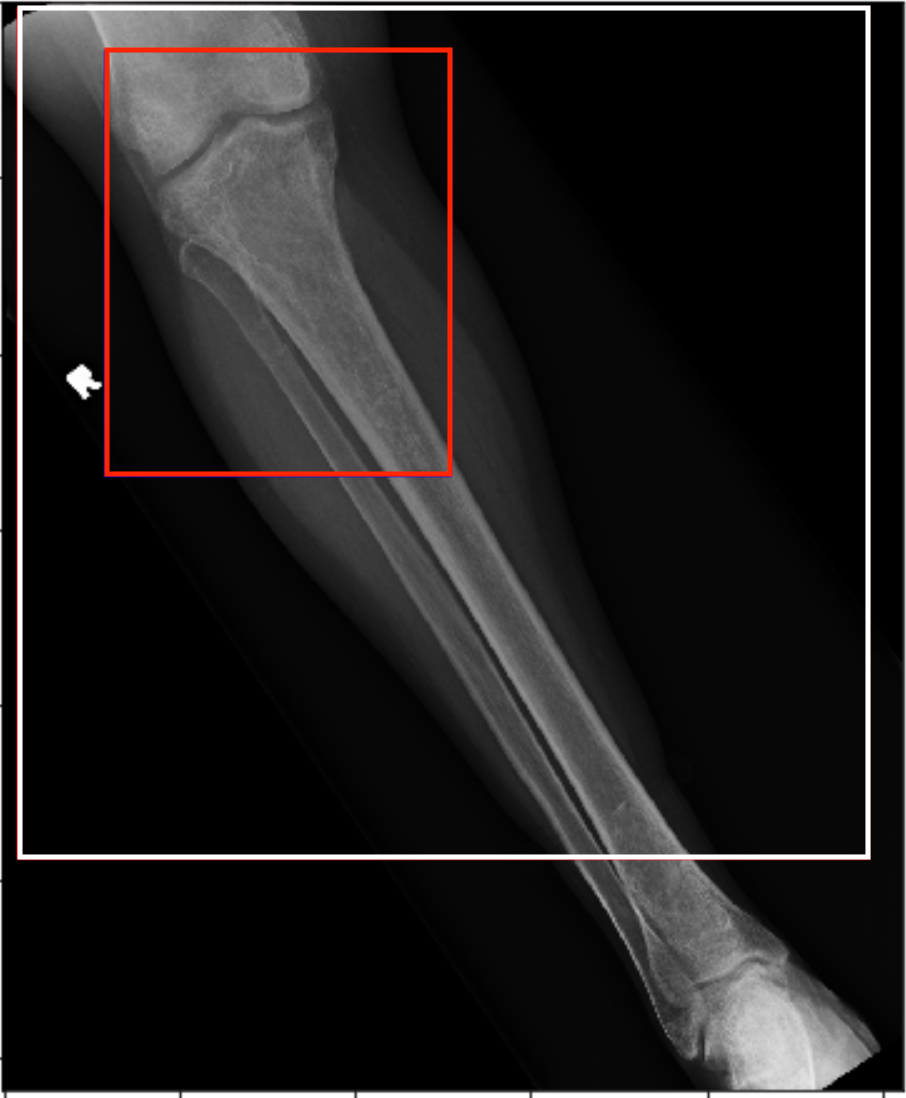}
      \includegraphics[width=2.5cm,height=3.1cm]{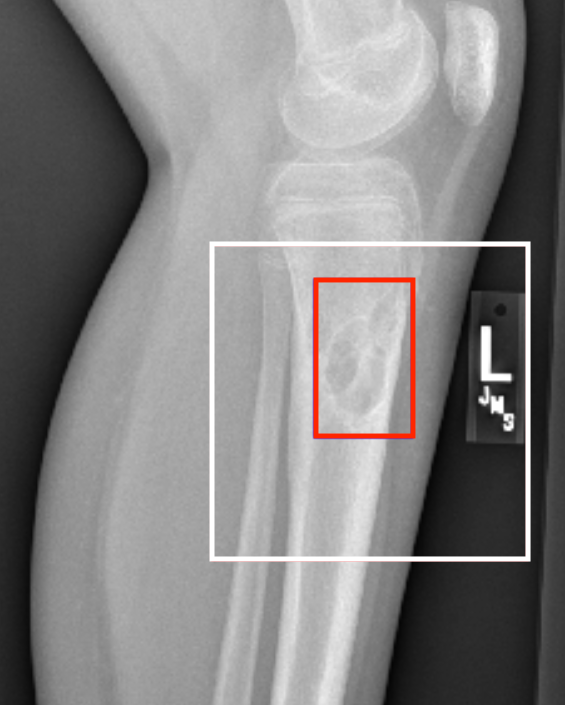}}
      \hspace{1mm}
    \subfigure[Femur]{\label{fig:femur}%
      \includegraphics[width=2.5cm,height=3.1cm]{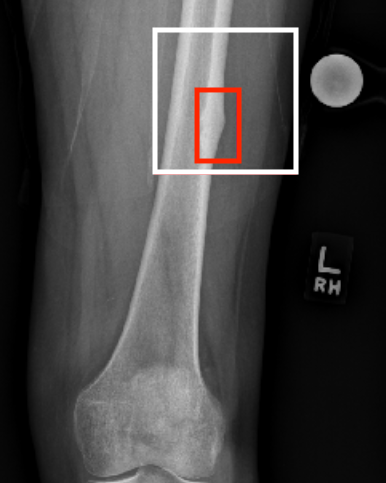}
      \includegraphics[width=1.8cm,height=3.1cm]{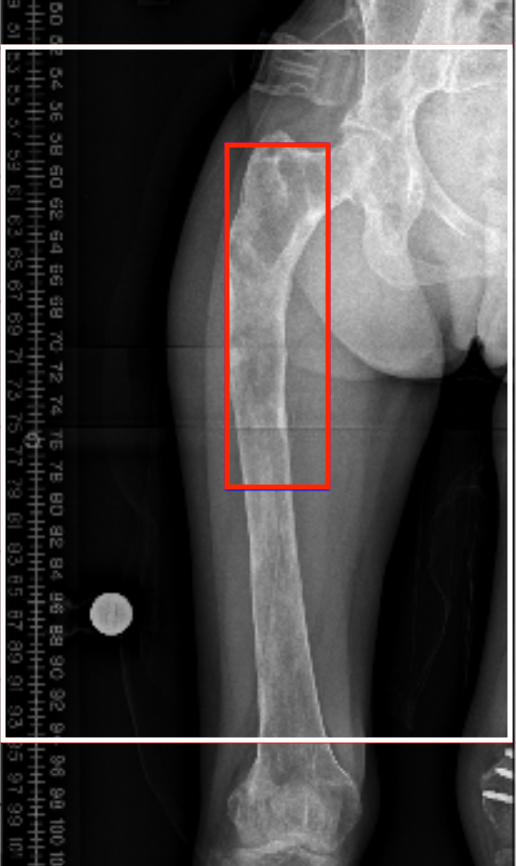}}
  \caption{\label{fig:patch}Bone lesion with expert-annotated bounding box (red). Random sized patches (white), cropped around lesions, used for training the generative model.}
\end{figure}

\section{Experimental Setup}
\subsection{Dataset}
A set of adult X-ray images showing bone anatomy with and without lesion are sourced from various U.S. hospitals and assessed by expert, board-certified radiologists by drawing bounding boxes around the target pathology (bone lesions) of concern (c.f. \figurename~\ref{fig:patch}). A test dataset is held out containing sufficient positive samples for evaluation and used at no point to train or fine-tune any model. The remaining dataset is then used for training and validating both the classifiers and the translation models.

\textbf{Classification Task:} Images with presence of confounding features (e.g., congenital deformity, fixation hardware) negatively impacted the model's classification performance. We thus removed those images from all datasets when training classification models. A summary of the data split, excluding augmented samples, is provided for the three investigated body parts in Table \ref{datasplit} (left). The generated images used for augmentation are only added to the training set but not the validation set.

\textbf{Translation Task:} 
We do not remove images from the lesion set when training the generative model, as it is trained on cropped image patches that are less affected by the confounding features. However, we remove images with confounding features from the non-lesion set to ensure that the augmented training set does not contain confounding images. The class split is kept balanced to facilitate training of the models. The images from the negative class in the training set that are not used to train the generative model are used for creating the augmented training set. Image patches are cropped from those images as described in Section~\ref{subsec:patches}. Table \ref{datasplit} (right) reports the distribution of the patch dataset and the configuration settings.

\begin{table}
    \caption{Datasets for each model (ratio denotes lesion:non-lesion class split). Left: images used for classification. Right: Extracted patches used for generation. Source samples are only non-lesion and used for creating the augmented sets. $s$ is the factor by which the patch is larger than the larger side of the bounding box. $n$ is the number of non-lesion images chosen against each lesion image.}
    \label{datasplit}
    \begin{minipage}{.5\linewidth}
      \centering
      \small
        \begin{tabular}{lrrr}
            \\
            \toprule
            \multicolumn{4}{c}{\bf Classification Task}\\
            \bf Body part & \bf Train & \bf Val & \bf Test \\
            \midrule
            Humerus & 268:2295 & 41:305 & 50:500    \\
            Tibia & 214:14482 & 22:1628 & 50:500    \\
            Femur & 32:4558 & 14:573 & 50:500       \\
            \bottomrule
        \end{tabular}
    \end{minipage}%
    \quad
    \begin{minipage}{.4\linewidth}
      \centering
      \small 
        \begin{tabular}{lrrrr}
            \\
            \toprule
            \multicolumn{5}{c}{\bf Translation Task}\\
            \bf Body part & \bf Train & \bf Source & \bf $s$ & \bf $n$\\
            \midrule
            Humerus & 536:536 & 4643 & 2 & 10\\
            Tibia & 515:515 & 4680 & 1 & 7\\
            Femur &  285:285 & 9171 & 2 & 10\\
            \bottomrule
        \end{tabular}
    \end{minipage} 
\end{table}

\begin{figure}[t]
    \subfigure[]{
        \includegraphics[width=0.45\linewidth]{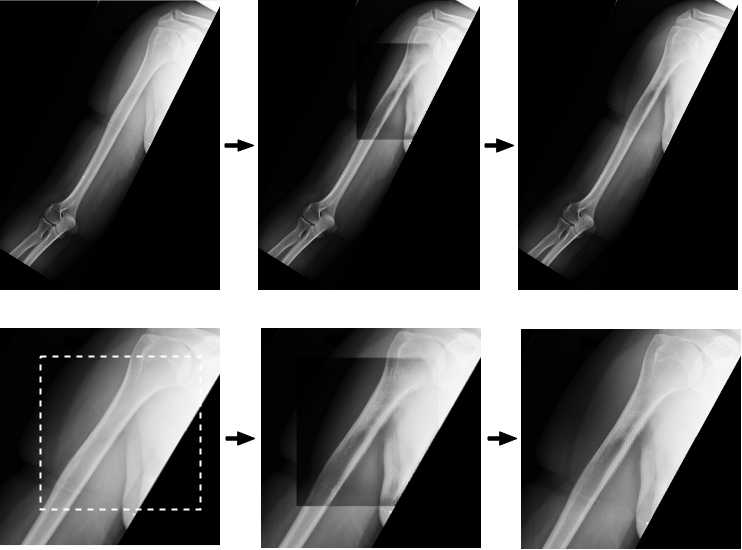}}
    \qquad
    \subfigure[]{
        \includegraphics[width=0.45\linewidth]{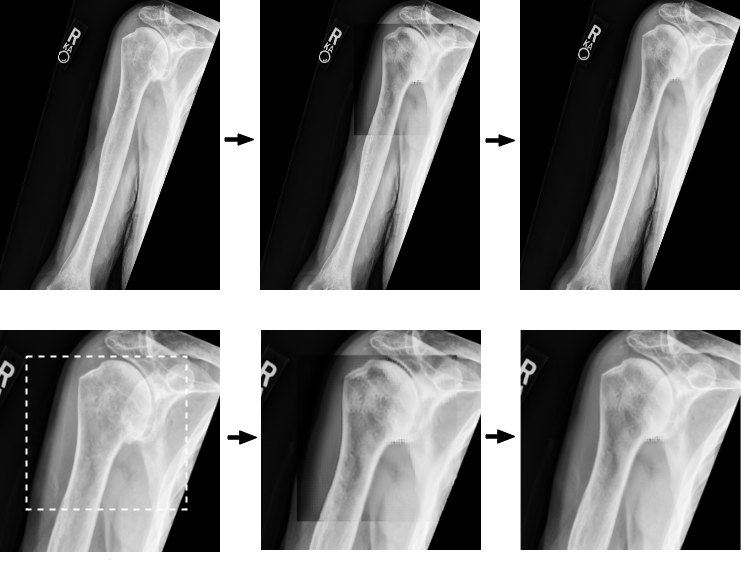}}
    \caption{Stages of patch translation for full-image (top) and selected patch, highlighted with white box (bottom): i) original, ii) translated and iii) blended.}
    \label{fig-blend}
\end{figure}

\subsection{Model Architecture and Optimization}
The classifier is a dilated residual net (DRN) \cite{yu2017dilated}. Dilated convolutional filters increase the receptive field of view and help capture finer details in high-resolution images. Images are downsampled to 1024x512 pixels in our experiments. To avoid overfitting on our comparatively small training set, our model was pretrained on a larger corpus of X-ray images for the auxiliary task of fracture detection. Training the classifier in this work involves fine-tuning of the last two convolutional blocks and the fully-connected layer of the model. Regularization is performed through augmentation procedures including linear transformations, along with weight decay. The model is optimized using Adam with an initial learning rate of 0.0001 which decays by a factor of 0.9 when the performance on the validation set plateaus.

The variability of the body part specific bone anatomy influenced our ability to train the translation model. Models on more diverse datasets like tibia could only be trained if the patch sizes were not larger than the bounding box ($s=1$). On the other hand, a comparably uniform anatomical view among humerus images allowed training with larger patch sizes ($s=2$). The adversarial loss weight influenced the qualitative results. Setting $\lambda_0=1$ resulted in a change in texture of the bone, rather than synthesis of a circular lesion. The default architecture and loss weighting as specified in \cite{liu2017unsupervised} proved to yield the best results. We found residual connections in the encoder and generator beneficial and hypothesize that copying the common features in the patch helps in training on such a small dataset. Figure \ref{fig-blend} demonstrates the blending process after translation using the default mask. 

\subsection{Transfer Learning}
\begin{figure}[t]
    \subfigure[Tibia]{
        \includegraphics[width=2.4cm,height=3.42cm]{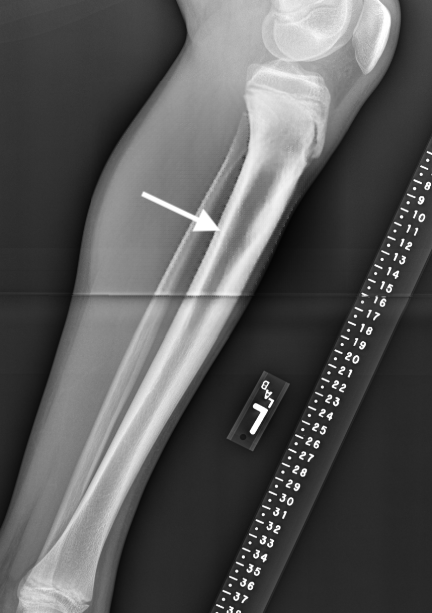}
        \includegraphics[width=2.4cm,height=3.42cm]{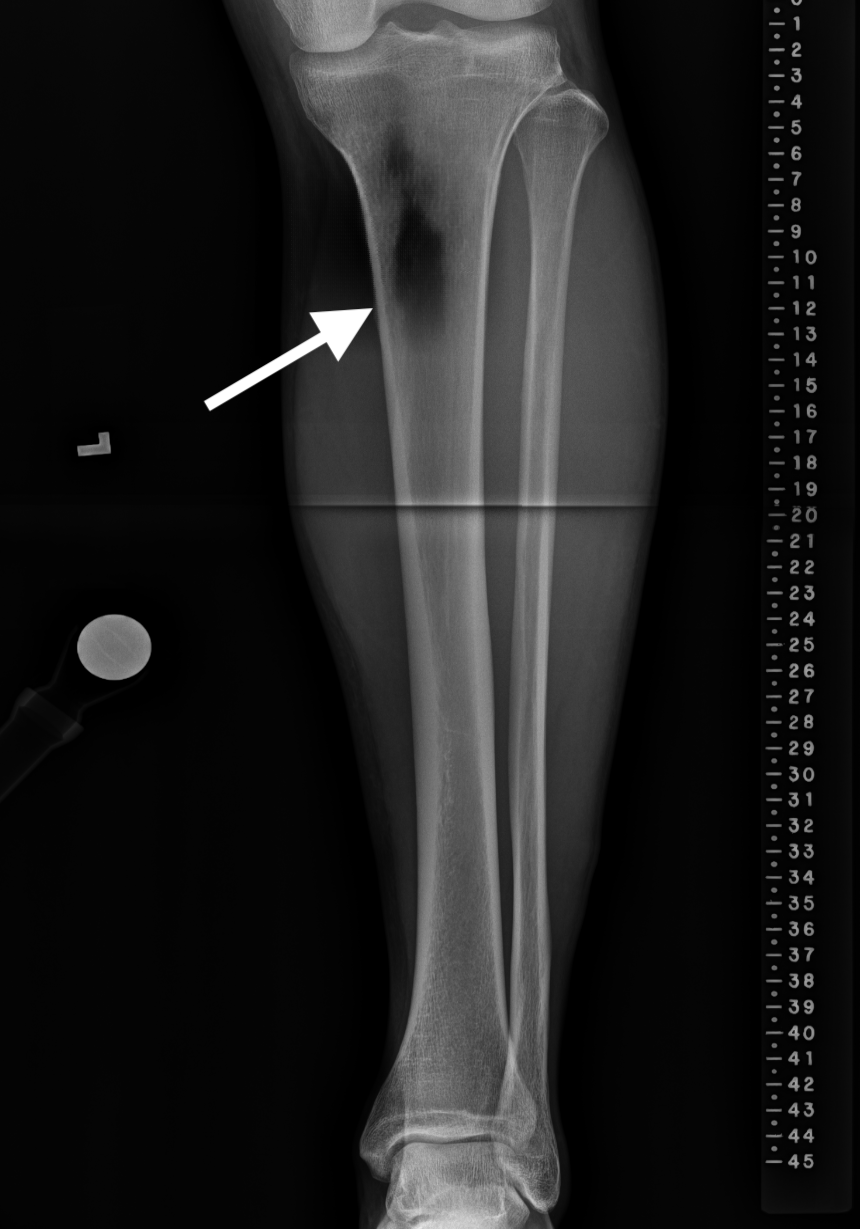}
        \includegraphics[width=2.4cm,height=3.42cm]{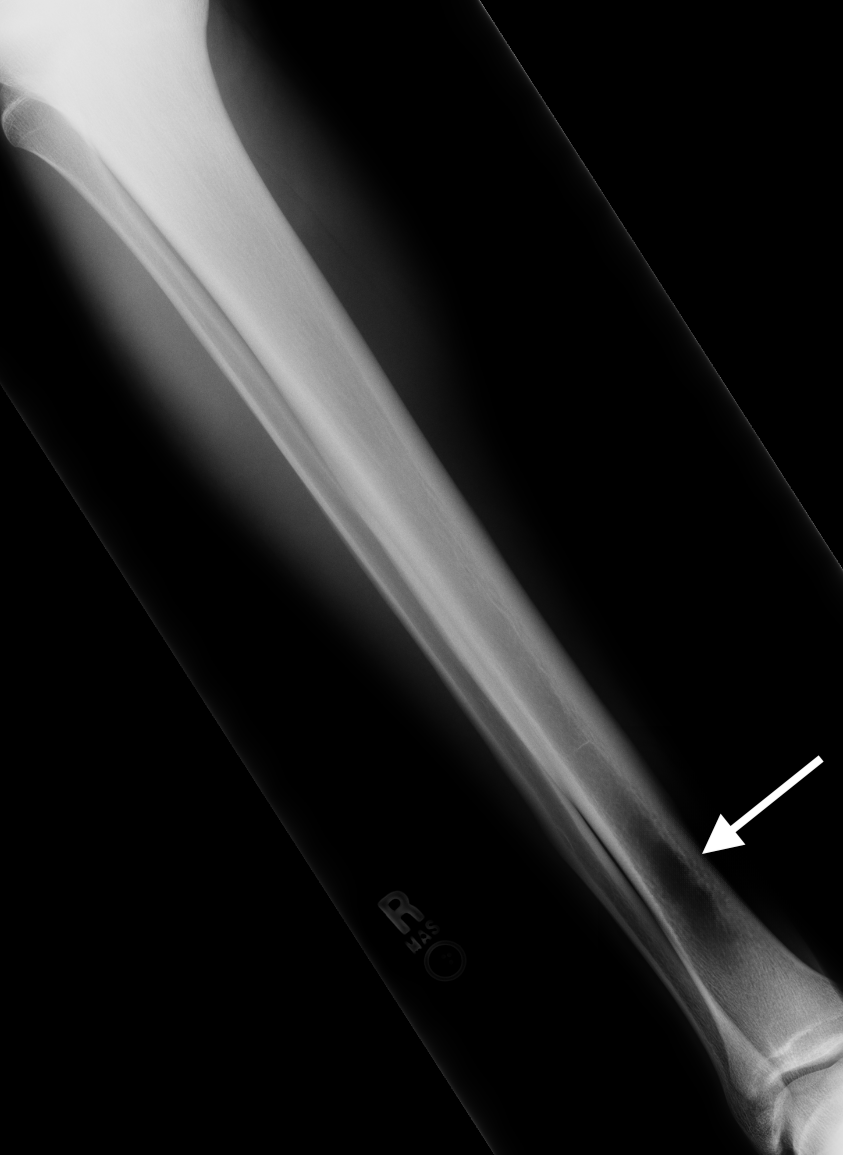}}
    \hspace{1mm}
    \subfigure[Femur]{
        \includegraphics[width=2.4cm,height=3.42cm]{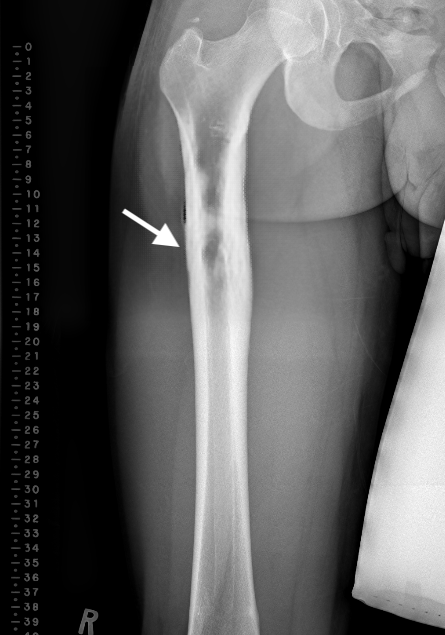}
        \includegraphics[width=2.4cm,height=3.42cm]{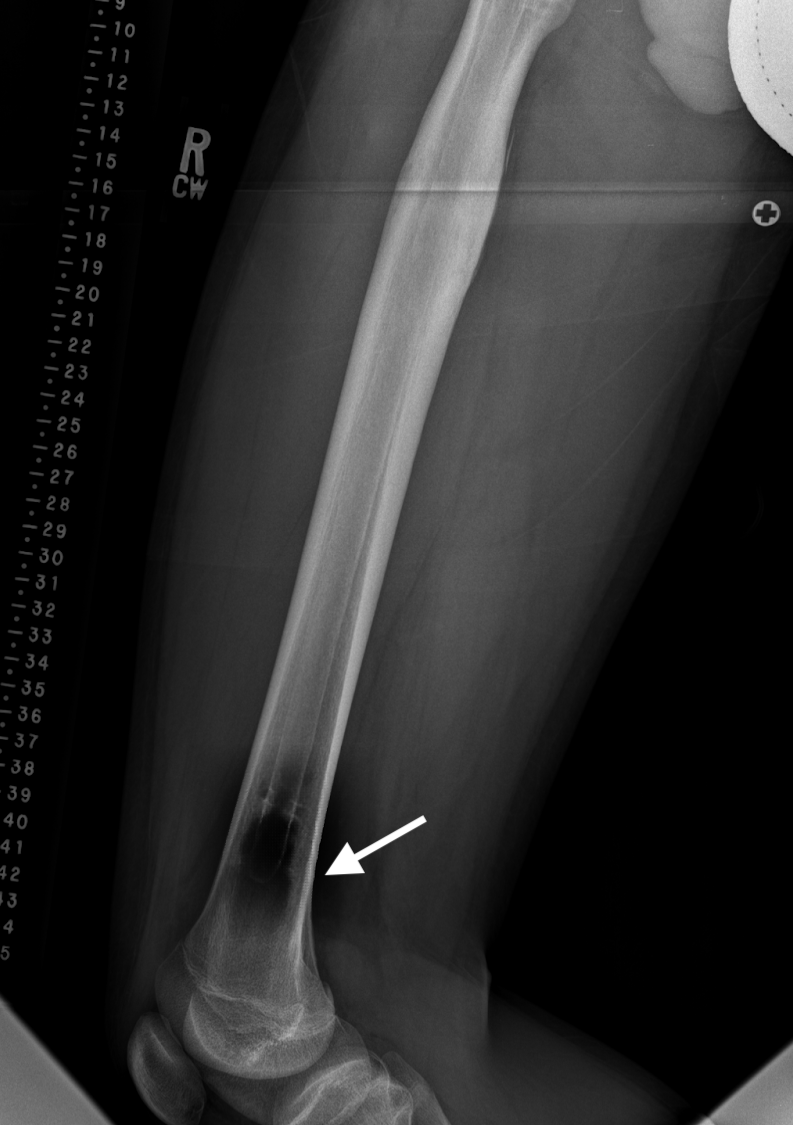}
        \includegraphics[width=2.4cm,height=3.42cm]{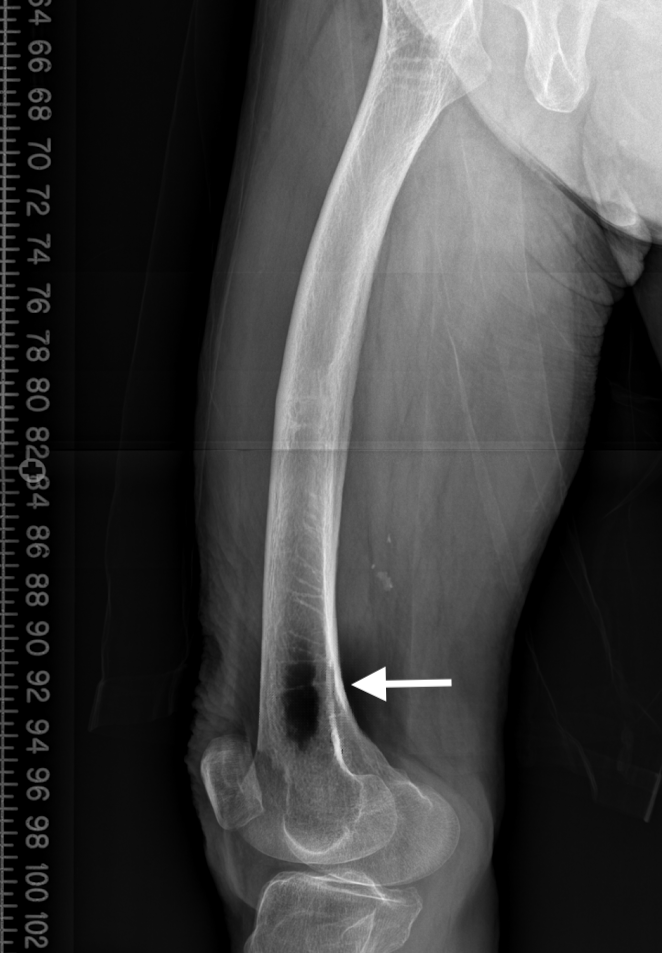}}
    \caption{\label{femur-tl}Bone lesion generated using transfer learning techniques. The variation in positional views within each body part makes it challenging to train a generative model.}
\end{figure}
In comparison to the available humerus X-rays, the available tibia and femur datasets were highly heterogeneous. We observed highly variable radiographic views and frequent confounding image content (e.g. external objects) in the not excluded positives. This made it particularly challenging to train a valuable generative model for tibia and unfeasible for femur, regardless of the patch size. We explored the potential of using transfer learning by i) employing the translation model trained on humerus to generate lesions on other body parts, ii) doing pseudo-labelling based on the humerus baseline classifier. For tibia we set $s=1$ to kept it consistent with the tibia-specific generative model. For femur we set $s=2$ to keep it consistent with the humerus configuration.

\subsection{Performance Measures}
We report the Area Under the ROC-Curve (AUC) and the bootstrapped 95\% Confidence Interval (CI). It was ensured that all models are compared on the same set of bootstrap samples. This allows us to examine the bootstrap-wise difference in AUC scores of models against the baseline. We consider a model to be significantly different to the baseline if the 95\% CI of those bootstrapped difference scores does not contain zero. We report Sensitivity (Sens) and Specificity (Spec) by defining an  Operating Point (OP) over the validation set as the point which minimizes $(1-\text{true positive rate})^2 + (\text{false positive rate})^2$ over the ROC curve. We focus on AUC scores since the operating point is, due to the low sample size of our validation set, highly variable and does not generalize well across experiments.

\begin{table}
    \caption{Ablation study of $t$ (threshold score of pseudo-labeller) reporting classifier performance on humerus test-set. Sensitivity and Specificity are calculated at the OP. Significantly different AUC with respect to baseline indicated with $^{*}$.}
    \label{threshold}
    \centering
    \small
    \begin{tabular}{lrrrrrrr}
        \toprule
        \bf Type & \bf $t$ & \bf Augmented Samples & \bf ROC AUC (CI 95\%) & \bf Sens. & \bf Spec. & \bf OP\\
        \midrule
         Baseline & 0 & 0 & 0.876 (0.817-0.926) & 0.9 & 0.776 & 0.455 \\
         \midrule
         Augmented & 0.70	& 1412	& 0.882 (0.829-0.928) & 0.80 	& 0.842 & 0.390 \\
         & 0.85	& 577	& 0.899 (0.854-0.939) & 0.82 	& 0.802 & 0.086 \\
         & 0.90	& 401	& \bf 0.924 (0.889-0.955)$^*$ & 0.84 	& 0.798 & 0.058 \\
         & 0.95	& 257	& 0.877 (0.820-0.926) & 0.90 	& 0.766 & 0.273 \\
        \bottomrule
    \end{tabular}
\end{table}
\begin{table}[t]
    \caption{Comparison of classifier model performance on tibia and femur test-sets. A translation model couldn't be trained for femur due to high diversity of radiographic view and insufficient samples. $\text{TL}_\text{G}$=Inference with humerus translation model, $\text{TL}_{\text{PL}}$=Pseudo-labelling with humerus baseline model.\label{tl-scores}}
    \subtable[Tibia]{
        \small
        \centering
        \begin{tabular}{lrrrrr}
            \toprule
            \bf Augment Type & \bf Augmented Samples & \bf ROC AUC (CI 95\%) & \bf Sens. & \bf Spec. & \bf OP \\
    		\midrule
    		Baseline & 0 &	0.618 (0.532-0.705) &  0.54 &	0.652 & 0.300 \\
            Augmented &	124 &	0.640 (0.547-0.732) &	0.6 &	0.542 & 0.244 \\
            $\text{TL}_\text{G}$ & 118 &	0.642 (0.550-0.735) &	0.52 &	0.66  & 0.290 \\
            $\text{TL}_\text{G} + \text{TL}_{\text{PL}}$ &	1264 &	\bf 0.698 (0.610-0.785)$^{*}$ &	0.74 &	0.464 & 0.066 \\
            \bottomrule
        \end{tabular}
    }\medbreak
    \subtable[Femur]{
        \small
        \centering
        \begin{tabular}{lrrrrrr}
            \toprule
            \bf Augment Type & \bf Augmented Samples & \bf ROC AUC (CI 95\%) & \bf 
            Sens. & \bf Spec. & \bf OP \\
            \midrule
    		Baseline  & 0 &	0.533 (0.441-0.627) &	0.64 &	0.376 &	0.010 \\
    		$\text{TL}_\text{G}$ & 579 & 0.601 (0.504-0.695)$^{*}$  &	0.56 &	0.61 &	0.012 \\
    		$\text{TL}_\text{G} + \text{TL}_{\text{PL}}$ & 1342 &	\bf 0.682 (0.594-0.764)$^{*}$ &	0.66 &	0.67 &	0.008 \\
            \bottomrule
        \end{tabular}
    }
\end{table}
\section{Results}
The augmentation set is composed of generated images that the baseline classifier assigns a confidence score of $t$ or higher. In the transfer learning setting, the humerus baseline classifier is used to select generated images for tibia and femur respectively. A grid search is performed on the validation set and $t$ is chosen to be the value that gives the highest validation set AUC ($t_\textrm{humerus}=0.9$, $t_\textrm{tibia}=0.9$, $t_\textrm{femur}=0.95$). To assess the influence of this parameter we report AUCs on the humerus test set for different values of $t$ in Table \ref{threshold}. We observe that the approach is sensitive to the choice of $t$ which, however, can be successfully chosen on the validation set. Adding either insufficient number of samples (larger $t$) or excessive low-quality samples (smaller $t$) reduces the benefit of data augmentation. We observed a significant increase in AUC of around 5\% over the humerus baseline model at $t=0.9$, as determined on the validation set.

For tibia we observed similar minor improvements ($\approx$2\%) when using either the humerus or tibia generative model. However, when further relying on the humerus baseline classifier for sample selection we observed a more substantial performance gain of around 8\% that was borderline to significant in the conducted test. For femur we observed significant gains in AUC when employing transferring knowledge from the humerus models. In particular, we observed an substantial improvement of around 15\% over the barely discriminative femur baseline classifier. See Table \ref{tl-scores} for the full quantitative analysis for tibia and femur when using transfer learning.
Figure \ref{femur-tl} illustrates some of the generated samples for tibia and femur obtained using transfer-learning based on the humerus model.

\section{Conclusion}
We trained a generative model that can represent some properties of the target pathology (bone lesions in X-ray) and synthesize those into sample patches drawn from another distribution (normal anatomy). When employing generative models for augmenting medical datasets, great care needs to be taken to avoid and control for possibly introduced bias. Future work should be concerned with the exploration of those limitations and explore the method's potential on both a more diverse set of disease pathology and other modalities.

\section*{Acknowledgements}
The project is funded by Imagen Technologies. The work presented in this manuscript is for research purposes only and is not for sale within the United States.

\bibliographystyle{unsrt}
\bibliography{references}
\end{document}